# Detection of Dataset Shifts in Learning-Enabled Cyber-Physical Systems using Variational Autoencoder for Regression

Feiyang Cai
*Vanderbilt University*
Nashville, TN
feiyang.cai@vanderbilt.edu

Ali I. Ozdagli
*Vanderbilt University*
Nashville, TN
ali.i.ozdagli@vanderbilt.edu

Xenofon Koutsoukos
*Vanderbilt University*
Nashville, TN
xenofon.koutsoukos@vanderbilt.edu

*Abstract*—Cyber-physical systems (CPSs) use learning-enabled components (LECs) extensively to cope with various complex tasks under high-uncertainty environments. However, the dataset shifts between the training and testing phase may lead the LECs to become ineffective to make large-error predictions, and further, compromise the safety of the overall system. In our paper, we first provide the formal definitions for different types of dataset shifts in learning-enabled CPS. Then, we propose an approach to detect the dataset shifts effectively for regression problems. Our approach is based on the inductive conformal anomaly detection and utilizes a variational autoencoder for regression model which enables the approach to take into consideration both LEC input and output for detecting dataset shifts. Additionally, in order to improve the robustness of detection, layer-wise relevance propagation (LRP) is incorporated into our approach. We demonstrate our approach by using an advanced emergency braking system implemented in an open-source simulator for self-driving cars. The evaluation results show that our approach can detect different types of dataset shifts with a small number of false alarms while the execution time is smaller than the sampling period of the system.

## I. Introduction

Recently, machine learning techniques, such as deep neural networks (DNNs), are extensively used in a wide variety of domains since they can handle complex tasks that cannot be easily solved by conventional techniques. On the other hand, cyber-physical systems (CPSs) are generally deployed in environments with high uncertainty and variability, which requires a high level of autonomy of the systems. There is no surprise that machine learning methods are increasingly used in CPSs to perform different difficult tasks, such as perception [1], planning [2], and control [3]. Although learning-enabled components (LECs) have demonstrated promising performance in such tasks in CPSs, the safety and reliability of LECs should be analyzed and ensured before deploying them to real-world systems, especially safety-critical systems. Unfortunately, the complex characteristics of the LECs can impede the analysis. Furthermore, LECs are typically trained using learning techniques such as supervised and reinforcement learning, and the implicit assumption for such learning techniques is the training and testing distribution are identical. However, although an LEC is trained extensively during the design time, the dataset shifts may still happen when it is applied to the real world. Dataset shifts may lead the LEC to be ineffective to predict large-error outputs, and further compromise the safety of the overall system. Therefore, the runtime detection method is very significant and necessary to guarantee the system safety and reliability of the system. The objective of the dataset shifts detection method is to quantify the degree of difference between the new test inputs and the training data and raise the alarm indicating the LEC may predict an erroneous output due to the dataset shifts.

Although many efforts have been made to detect the dataset shifts or out-of-distribution examples in neural networks, especially regarding classification tasks [4], [5], such techniques perform detection only using a single targeted example and may result in a large number of false alarms when they are directly applied to CPS due to the dynamical nature of CPS. Only recently, a new method based on inductive conformal anomaly detection (ICAD) [6] is proposed in [7], where the robustness of the detection is improved by using multiple examples sampled from the VAE model. In addition, most out-of-distribution detection methods utilize only the inputs but ignore the outputs of LECs. However, these methods may fail to detect some types of dataset shift, in particular, where the change in the distribution is observed for the output. In [8], extending the method in [7], an adversarial examples detection method for learning-enabled CPS is proposed by utilizing the VAE for regression model. By using such model, a VAE and a regression model can be trained jointly, which enables the detector to take the LEC output into account. In this paper, the VAE for regression model is also used, but we focus on the detection of a variety of dataset shifts. Moreover, we

The material presented in this paper is based upon work supported by the National Science Foundation (NSF) under grant number CNS 1739328, and the Defense Advanced Research Projects Agency (DARPA) through contract number FA8750-18-C-0089. The views and conclusions contained herein are those of the authors and should not be interpreted as necessarily representing the official policies or endorsements, either expressed or implied, of NSF, or DARPA.

incorporate layer-wise relevance propagation (LRP) into the nonconformity measure to improve the robustness of detection.

The first contribution of this paper is formal definitions of a variety of dataset shifts in learning-enabled CPS. We introduce three different types of dataset shifts in this paper, and provide typical examples for each type of dataset shift aiming at a realistic CPS – advanced emergency braking system (AEBS).

Our second contribution is an approach for detecting a variety of dataset shifts in learning-enabled CPS. A VAE for regression model is employed in our approach in order to take into consideration both the input and output of the the regression LEC. Additionally, the layer-wise relevance (LRP) algorithm is incorporated into the detection algorithm to further improve the robustness of the detection. The main benefit of this method is to treat the input features differently depending on their influence on the output of LEC.

The final contribution is the comprehensive evaluation using a simulation case study – AEBS, which is implemented in CARLA [9], an open-source self-driving simulator. In AEBS, our evaluation focuses on the perception LEC, which estimates the distance from the host vehicle to the obstacle. For different types of dataset shifts, we design different experiments to evaluate our approach. The experimental results demonstrate the proposed approach can detect different types of dataset shifts with a very small number of false alarms.

## II. DATASET SHIFTS IN LEARNING-ENABLED CYBER-PHYSICAL SYSTEMS

Supervised learning algorithms are deployed with the assumption that the training and testing samples are drawn from the same distribution. However, when a change happens in the distribution of the testing data, the learning model most likely becomes ineffective for predicting the outcome for the new test data. A change in the distribution of features for the unseen data is described as dataset shift. It is crucial to analyze and formalize this phenomenon such that the learning-enabled models can be improved.

This study focuses on the detection of a variety of dataset shifts. Therefore, in this section, we first introduce the formal definitions of datashift sets we selected for this study. Next, we discuss the AEBS architecture briefly. We also presented some and provided palpable datashift examples relevant to AEBS. Lastly, we presented the problem formulation explaining the dataset shifts detection problem.

### A. Formal Definitions of Dataset Shifts

*1) Covariate shift:* Covariate shift is one of the most basic and common dataset shifts observed in real-life [10]. Suppose that, we have a supervised learning model that is trained with an input-output pair such that $T = \{(x_i, y_i)\}_{i=1}^n$. The general assumption for this model is that the input, $x$ is IID and drawn from $P_{\text{train}}(x)$. This model can be used for making predictions based on the conditional probability, $P(y|x)$ for some $y$ given new $x$. Covariate shift usually occurs when the distribution of the input, $P(x)$ changes after training while the conditional probability $P(y|x)$ remains the same. More formally, $P_{\text{train}}(x) \neq P_{\text{test}}(x)$ and $P_{\text{train}}(y|x) = P_{\text{test}}(y|x)$.

*2) Target shift:* Target shift is the reverse covariate shift where the distribution over $y$, $P(y)$ changes but the output-conditional distribution, $P(x|y)$ remains the same. A target shift can affect the prediction accuracy significantly [11]. The formal definition is that the training and testing distribution for output variables changes such that $P_{\text{train}}(y) \neq P_{\text{test}}(y)$ whereas the conditional probability of $x$ given $y$ remains unchanged, i.e. $P_{\text{train}}(x|y) = P_{\text{test}}(x|y)$.

*3) Label concept shift:* A concept shift is simply a contextual shift where the underlying relationship between input and output changes while the distribution over input is preserved [12]. In concept shift, we assume $P_{\text{train}}(x) = P_{\text{test}}(x)$ while $P_{\text{train}}(y|x) \neq P_{\text{test}}(y|x)$. For label concept shift, instead of input variables, the distribution over output stays the same such that $P_{\text{train}}(y) = P_{\text{test}}(y)$.

### B. AEBS Architecture

The use of LECs becomes a popular choice in many classes of CPS for increasing the level of autonomy of the system. A quintessential example of learning-enabled CPS – advanced emergency braking system (AEBS), is proposed in [7], which is an automobile system designed to detect the approaching obstacle and stop the host vehicle safely. A simplified system architecture is presented in Fig. 1. There are two LECs in this architecture: a perception LEC tries to estimate the distance to the approaching obstacle using the images captured by a camera, which is trained by supervised learning; a control LEC consumes the estimated distance and the velocity of the host vehicle, and generates a braking force to safely stop the host vehicle, which is trained by reinforcement learning. More details about the AEBS can be found in [7].

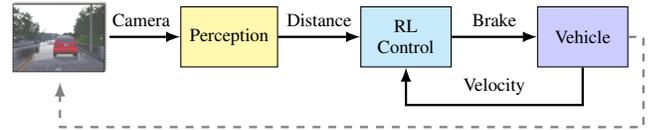

Fig. 1. Advanced emergency braking system architecture [7].

### C. Examples to Dataset Shifts in AEBS

Braking systems are occasionally subjected to dataset shifts, in particular, covariate shifts. Suppose that we have a learning-enabled system that predicts the braking distance, $y$ for a given camera input, $x$. An AEBS trained with camera images taken during the daylight may predict the braking distance incorrectly for night time. For this system, the probability distribution of the daylight images, $P_{\text{day}}(x)$ diverges from the night time images, $P_{\text{night}}(x)$. However, in reality, the braking distance should be independent of the time of the day and the resulting light conditions. Thus, the conditional probability $P(y|x)$ should not change.

AEBS may be also susceptible to the target shift. For typical training scenarios, we assume the probability distribution for the braking distance, $P_{\text{train}}(y)$ is uniform. In real-life situations, unique traffic patterns may impose a distribution, $P_{\text{test}}(y)$ that does not match this assumption. We hypothesize that the shift in $P(y)$ may be critical for proper operations of AEBS and requires attention.

For AEBS, the braking distance prediction model, $P(y|x)$ assumes that the vehicle types and shapes conform to the specifications imposed by the Department of Transportation. In the future, the specifications may change drastically as a response to the safety criteria of progressively developing autonomous vehicle technologies. While the braking distance distribution, $P(y)$ remains the same, due to the safety-related specification changes, the prediction model may not be able to correctly estimate the safe braking distance. This can be solved by retraining the model. The scope of this paper focuses on the detection of the label concept shift.

### D. Problem Formulation

Learning techniques, such as supervised and reinforcement learning, are generally used to design the LECs. The implicit assumption for such learning techniques is that the training and testing distribution are identical. Although the LECs are successfully trained and the training errors are satisfactory during design time, the LECs may still become ineffective due to the *dataset shifts*, where the joint distribution of inputs and outputs differs between training and test stages. Dataset shifts may lead the LEC to make erroneous predictions and undermine the safety of the system.

The problem considered in our paper is to robustly detect a variety of dataset shifts in learning-enabled CPS in real-time. The LEC receives the inputs one by one and predicts the outputs during the system operation. The objective is to compute a measure quantifying the degree to which a dataset shift has happened.

Online detection algorithms must be robust with a small number of false alarms. Detection of dataset shifts should take not only input but also output into consideration since change of the distribution of output can also lead to dataset shifts.

### III. Variational Autoencoder for Regression

A variational autoencoder (VAE) is a generative model whose objective is to learn an underlying probability distribution over the high-dimensional input data points. Similar to an autoencoder, a VAE models a relationship between the high-dimensional input data point and its low-dimensional latent representation. In addition, the latent space is regularized during the training using a probabilistic manner, which make the latent encodings to have capacity of generating of new data [13]. Furthermore, a VAE for regression model is proposed in [14] trying to learn a conditional latent representation on a specific regression target variable. Fig. 2 presents the architecture of this VAE for regression model.

The core idea of the VAE for regression model is, the latent representation $z$ is conditioned on the target variable $c$ predicted by a regression network, and therefore, the latent distribution is represented by a conditional Gaussian distribution $p(z|c)$ in lieu of a single Gaussian prior $p(z)$ used in traditional VAE. There are two additional components in VAE for regression model comparing with VAE: *regressor* and *latent generator*. The regressor employs a regular regression network $q(c|x)$ with an additional output to infer the target variable $c$ and its uncertainty, and the latent generator feeds the target variable $c$ into the latent space trying to condition the latent distribution on this predicted variable. The loss function of the VAE for regression model can be expressed as

$$\begin{aligned}\mathcal{L}(\theta,\phi_c,\phi_z;x) = &- D_{\text{KL}}(q_{\phi_c}(c|x)||p(c)) \\ &+ \mathbb{E}_{z\sim q_{\phi_z}(z|x)}[\log p_\theta(x|z)] \\ &- \mathbb{E}_{c\sim q_{\phi_c}(c|x)}[D_{\text{KL}}(q_{\phi_z}(z|x)||p(z|c))].\end{aligned}$$

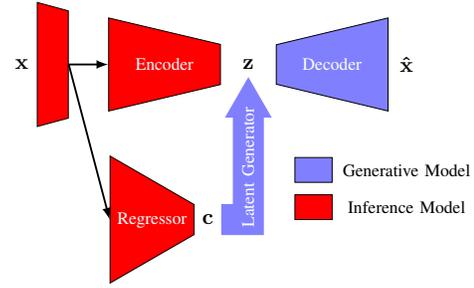

Fig. 2. VAE for regression model [14].

The first term corresponds to the regressor whose objective is to regularize the distribution of prediction variable $c$ with the ground-truth prior $p(c)$. Similar to a traditional VAE, the second term encourages the decoder to reconstruct the input from the latent representations with a small reconstruction error. The third term tries to minimize the KL divergence between the approximate posterior and the regression specific prior $p(z|c)$. Using such model and loss function, we can train the VAE and the regression model in a single network.

### IV. Detection of Dataset Shifts

#### A. Inductive Conformal Anomaly Detection

Our approach uses the inductive conformal anomaly detection (ICAD) as the basic framework which requires a suitable *nonconformity measure* defined to quantify how different a test data is relative to the training dataset. The ICAD algorithm can be split into offline and online phases. During the offline phase, given a training dataset $\{(x_1, y_1), \ldots, (x_l, y_l)\}$, where each example $(x_i, y_i) : i = 1, \ldots, l$ consists of observations $x \in X$ and labels $y \in Y$, the first step is to split this training dataset into a *proper training set* $\{(x_1, y_1), \ldots, (x_m, y_m)\}$ and a *calibration set* $\{(x_{m+1}, y_{m+1}), \ldots, (x_l, y_l)\}$. Then, for each calibration data, we compute the nonconformity score with respect to the proper training set by

$$\alpha_i = A\Big(\{(x_1,y_1),\ldots,(x_m,y_m)\},(x_i,y_i)\Big), i = m+1,\ldots,l.$$

All calibration nonconformity scores are sorted for the online detection phase. At runtime, for a test example $(x_{l+1}, y_{l+1})$, we compute its nonconformity score $\alpha_{l+1}$ using the nonconformity measure $A$ with respect to the proper training set. Then, the $p$-value for the test example is defined as the fraction

of calibration examples that equally or more nonconforming than this test example, and can be computed as

$$p_{l+1} = \frac{|\{i = m+1, \ldots, l\} \,|\, \alpha_i \geq \alpha_{l+1}|}{l - m}. \quad (1)$$

In ICAD method, for a test example, if the $p$-value is smaller than a threshold $\epsilon \in (0, 1)$, it will be recognized as a conformal anomaly. However, it is not robust to detect the anomaly only using a single example. In [15], it is pointed that, if a sequence of data points are in the same distribution as the proper training set, the corresponding $p$-values are independent and uniformly distributed between 0 to 1. Therefore, out-of-distribution detection can be performed by testing the hypothesis that the sequence of $p$-values are independent and uniformly distributed between 0 to 1. Furthermore, we can use martingale to test this hypothesis [16]. *Simple mixture martingale* is a typical martingale and is defined as

$$M_N = \int_0^1 \prod_{i=1}^N \epsilon p_i^{\epsilon-1} d\epsilon,$$

where $N$ is the size of the sequence. Such martingale will grow only if there are many $p$-values in the sequence indicating there are unusual examples.

### B. Nonconformity Measure with Layer-wise Relevance Propagation

In order to scale the ICAD method to the high-dimensional inputs, variational autoencoder (VAE) is proposed to compute the nonconformity scores efficiently [7]. The mean square error between the input $x$ and the reconstructed output $\hat{x}$ is utilized as the nonconformity measure. Although the evaluation in [7] shows a promising result, there are still some defects of VAE-based nonconformity measure. It is possible that some nonconformal features of the input rarely or do not contribute to the LEC prediction. As an illustrative example, if the VAE has difficulty generating fine-granularity details of the original input image, the VAE-based nonconformity measure will result in a large nonconformity score. Therefore, it is unfair to treat all features equally for nonconformity measure when the features contribute to the LEC prediction differently. We incorporate the *layer-wise relevance propagation* (LRP) algorithm into the VAE-based nonconformity measure to compensate for such defects and improve the robustness of the detector.

LRP is normally used as a tool for interpreting neural networks by identifying which input features contribute most to the LEC predictions [17]. We focus on neural networks with image input in this paper. Considering an input image $x$ is of size $H \times W \times C$ where $H$, $W$, and $C$ denote the height, width, and channel respectively. We denote each input element or pixel as $x^{h,w,c}$ where $h \in \{0, \ldots, H-1\}, w \in \{0, \ldots, W-1\}$ and $c \in \{0, \ldots, C-1\}$. LRP computes the relevance $r^c$ which has the same size as the input $x$ by running a backward pass in the monitored LEC. In addition, we convert $r^c$ to a grayscale relevance map $r$ to use the relevance in the nonconformity measure. To be brief, a function $r = G(x_0; f)$ is defined to represent the LRP procedure generating a grayscale relevance map $r$ for the monitored LEC $f$ and a given input $x_0$.

In our approach, we use this relevance to weight the contribution to the nonconformity score, and further, we can define the nonconformity measure with LRP by weighting the reconstruction error of VAE using the relevance map $r$

$$\begin{aligned}\alpha &= A_{\text{VAE-LRP}}(x, \hat{x}, r) \\ &= \frac{1}{H \times W \times C} \sum_{h=0}^{H-1} \sum_{w=0}^{W-1} \sum_{c=0}^{C-1} r^{h,w}(x^{h,w,c} - \hat{x}^{h,w,c})^2. \end{aligned} \quad (2)$$

Therefore, the nonconformity measure treats input features differently based on their relevance or influence on the output of the monitored LEC.

### C. Detection Method

The main idea of the VAE-based approach is to generate multiple IID examples similar to the input from the learned probability distribution of the latent space. The computed $p$-values for these generated samples are independent and uniformly distributed in $[0, 1]$. Then, as described before, the martingale test can be employed to test if any dataset shifts happen in the distribution of the testing data relative to the training dataset. In this paper, the VAE for regression model is used to generate multiple IID examples in order to take the output into account for detecting dataset shifts. Besides, the nonconformity scores are computed by the VAE-based nonconformity measure with LRP (Eq. 2). The relevance between the input features and the output can be computed using the LRP algorithm by propagating the prediction output through the regressor of the VAE for regression model. The detailed procedures of the algorithm are described below.

*1) Offline training:* Our approach is based on ICAD, therefore, the first step is to split the training set $\{(x_1, y_1), \ldots, (x_l, y_l)\}$ into a proper training set $\{(x_1, y_1), \ldots, (x_m, y_m)\}$ and a calibration set $\{(x_{m+1}, y_{m+1}), \ldots, (x_l, y_l)\}$. Then, we train a VAE for regression model $(y, \hat{x}) = f(x)$ using the proper training set. The model performs two tasks: regression task which is defined by a mapping from the input $x$ to the target variable $y$; generation task which is defined by a mapping from the input $x$ to the reconstructed output $\hat{x}$. For each example $x_j : j \in \{m+1, \ldots, l\}$ in the calibration set, a single reconstructed input $\hat{x}_j$ is sampled and generated from the latent space, and the relevance $r_j$ between the input features and the prediction output can be computed by the LRP algorithm ($r_j = G(x_j; f)$). Incorporating the relevance $r_j$, the nonconformity score $\alpha_j$ can be computed by the nonconfromity measure $A_{\text{VAE-LRP}}$ defined in Eq. (2). The precomputed nonconformity scores for data in calibration set are sorted and stored to be used for online detection.

*2) Online detection:* During runtime, the test inputs $(x'_1, \ldots, x'_t, \ldots)$ arrives to the LEC one-by-one. Based on the idea in [7], for each test input $x'_t$, $N$ examples $\hat{x}'_{t,1}, \ldots, \hat{x}'_{t,N}$ are generated from the learned posterior distribution in the latent space. In addition, a relevance map $r'_t$ is computed by $G(x'; f)$. Then, for each generated example $\hat{x}'_{t,k}$, the nonconformity score $\alpha'_{t,k}$ and its $p$-value $p_{t,k}$ are computed using Eq. (2) and Eq. (1). Since the generated examples are IID, the $p$-values $(p_{t,1}, \ldots, p_{t,N})$ are independent and uniformly distributed in $[0, 1]$. A martingale test can be applied for every new input example $x'_t$ at time $t$

$$M_t = \int_0^1 M_t^\epsilon d\epsilon = \int_0^1 \prod_{k=1}^N \epsilon p_{t,k}^{\epsilon-1} d\epsilon. \quad (3)$$

If the dataset shift happens, the martingale $M_t$ will grow up due to many small $p$-values in the sequence. In addition, as described in [7], a stateful CUSUM detector $S$ is employed to generate alarms when the martingale becomes consistently large. Algorithm 1 summarizes the proposed method.

---

**Algorithm 1** Dataset shifts detection using VAE for regression with LRP

---

**Input:** Input training set $\{(x_1, y_1), \ldots, (x_l, y_l)\}$, input sequence $(x'_1, \ldots, x'_t, \ldots)$, number of calibration examples $l - m$, number of examples $N$ generated by the VAE, threshold $\tau$ and parameter $\delta$ of CUSUM detector
**Output:** Output boolean variable $Anom_t$
**Offline:**
1: Split the training set $\{(x_1, y_1), \ldots, (x_l, y_l)\}$ into the proper training set $\{(x_1, y_1), \ldots, (x_m, y_m)\}$ and calibration set $\{(x_{m+1}, y_{m+1}), \ldots, (x_l, y_l)\}$
2: Train a VAE for regression $f$ using the proper training set
3: **for** $j = m + 1$ to $l$ **do**
4:    Generate $\hat{x}_j$ using the trained VAE for regression
5:    Compute the relevance map $r_j = G(x_j; f)$
6:    $\alpha_j^\Gamma = A_{\text{VAE-LRP}}(x_j, \hat{x}_j, r_j)$
7: **end for**
8: $\{\alpha_{m+1}, \ldots, \alpha_l\} = \text{sort}(\{\alpha_{m+1}^\Gamma, \ldots, \alpha_l^\Gamma\})$
**Online:**
9: **for** $t = 1, 2, \ldots$ **do**
10:    Compute the relevance map $r'_t = G(x'_t; f)$
11:    **for** $k = 1$ to $N$ **do**
12:      Generate $\hat{x}'_{t,k}$ using the trained VAE for regression
13:      $\alpha'_{t,k} = A_{\text{VAE-LRP}}(x'_t, \hat{x}'_{t,k}, r'_t)$
14:      $p_{t,k} = \frac{|\{i=m+1,\ldots,l\} \mid \alpha_i \geq \alpha'_{t,k}|}{l-m}$
15:    **end for**
16:    $M_t = \int_0^1 \prod_{k=1}^N \epsilon p_{t,k}^{\epsilon-1} d\epsilon$
17:    **if** $t = 1$ **then**
18:      $S_t = 0$
19:    **else**
20:      $S_t = \max(0, S_{t-1} + M_{t-1} - \delta)$
21:    **end if**
22:    $Anom_t \leftarrow S_t > \tau$
23: **end for**

---

## V. EVALUATION

We demonstrate our approach using a simulation case study – advanced emergency braking system (AEBS), which is implemented in CARLA [9]. We conduct all experiments on a 6-core Ryzen 5 desktop with a single GTX 1080Ti GPU.

### A. Experimental setup

As mentioned in Sec.II-B, our evaluation focuses on the perception LEC whose task is to estimate the distance to the nearest front obstacle using the images captured by the onboard camera. For collecting the training dataset, we vary the initial distance to the obstacle $d_0$ and initial velocity $v_0$ of the host vehicle. In addition, we control the precipitation parameter for the training dataset which is randomly sampled from the interval $[0, 20]$. The proper training set for the VAE for regression model training consists of 15920 images. Furthermore, using the simulations with the same settings, we collect additional 3980 images for the calibration dataset. We should note that the probability distribution of the ground-truth distance, $P_{\text{train}}(y)$ is nearly uniformly distributed in the range $[0\,\text{m}, 50\,\text{m}]$ so that the training dataset is almost balanced.

The VAE for regression model can be implemented by a convolutional neural network (CNN), and we use the same network architecture and training settings as [8]. After the training phase, the training and testing errors (mean absolute error) of the regressor are $0.04\,\text{m}$ and $0.06\,\text{m}$ respectively. In addition, by applying t-distributed stochastic neighbor embedding (t-SNE) [18], the encodings in the latent space are projected into low-dimensional representations which are plotted in Fig. 3. The plot shows that the model is able to disentangle the distance-related dimension from the latent space.

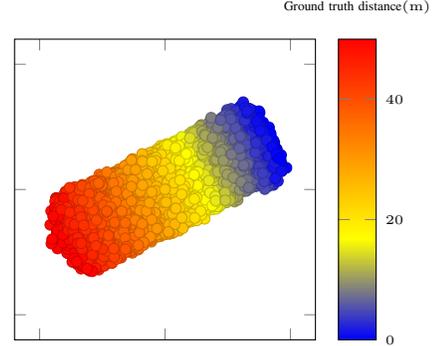

Fig. 3. 2-dimensional latent representations estimated by VAE for regression.

The nonconformity scores for data in the calibration set are precomputed and sorted for the online detection. For each test example during the online phase, the VAE for regression model generates $N = 10$ examples used for detection. We illustrate our approach using an in-distribution episode firstly and plot the absolute prediction error between the ground-truth and predicted distance to the obstacle, the $p$-value, and the output of the detector $S$ computed using the logarithm of martingale $M_t$ and $\delta = 4$ in Fig. 4. The results using nonconformity measures with and without the LRP algorithms are both plotted for comparing, which are denoted by (VAE-R) and (VAE-R, LRP) respectively in the plots. The results of the in-distribution episode show that, for nonconformity measure without the LRP algorithm, the $p$-values are almost randomly distributed in $[0, 1]$, and thus, the detector stays in a low value indicating there is no dataset shift happens. As for the nonconformity measure with the LRP algorithm, the $p$-values are randomly distributed between 0 to 1 at first, but decrease at the end of the episode. The reason for this phenomenon is, near the end of the episode, the lead vehicle occupies more pixels in the image, more pixels are very relevant to the LEC output, and the nonconformity scores are computed by taking more pixels into account. Although the $p$-values decrease a little, the detector $S$ is still smaller than the threshold 40 indicating there is no dataset shift during the episode.

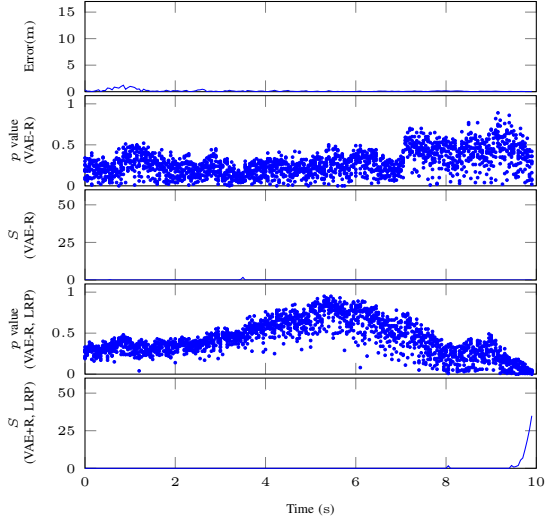

Fig. 4. Episode with in-distribution inputs.

### B. Covariate shift

In AEBS, we can control the precipitation parameter to enforce a testing distribution for input different from the training dataset. In the training dataset, the precipitation parameter is randomly sampled from 0 to 20. As for testing, we generate 50 episodes where the precipitation parameter is randomly sampled from 30 to 100.

A covariate shift episode is shown in Fig. 5. The error of the perception LEC is palpably larger than the error for in-distribution data and it can exceed $15\,\text{m}$. The $p$-values come down to almost 0 and the detector indicates the dataset shift happens. We also evaluate our approach using the 50 in-distribution episodes (precipitation parameter: $[0, 20]$) and 50 covariate shift episodes (precipitation parameter: $[30, 100]$) by considering different values of $N$. Table I reports the false alarms for detection covariate shift using the VAE-based nonconformity measure with and without the LRP algorithm. The results show that our approach can detect covariate shift with a very small number of false alarms.

### C. Target shift

For this shift, we consider a casual scenario, where unique traffic patterns impose specific braking distance distributions. More specifically, we assume all data between $15\,\text{m}$ to $45\,\text{m}$ are not included in the training dataset. An episode for the target shift is shown in Fig. 6, which contains the data between $0\,\text{m}$ to $50\,\text{m}$. In this case, the neural network does not work properly, and the results show the neural network predicts the outputs with large errors. The $p$-values come to nearly zero and the detector grows up, which indicates the dataset shift happens during the episode. Besides, Table I also reports the false alarms for detection such target shift, and the results demonstrate our approach can detect the target shift with a very small number of false alarms.

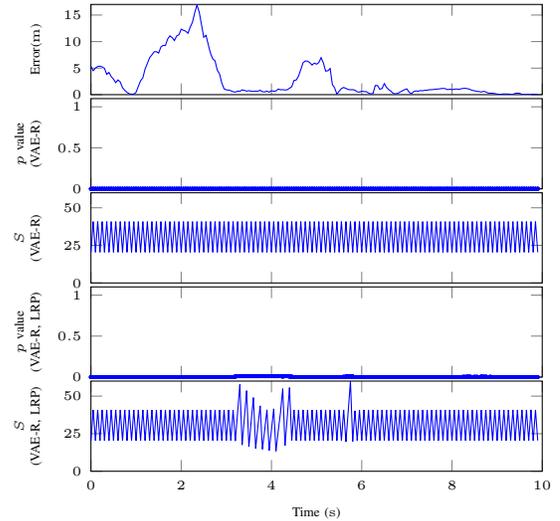

Fig. 5. Episode with dataset shift (covariate shift).

TABLE I
FALSE ALARMS FOR THE ADVANCED EMERGENCY BRAKING SYSTEM.

| Types | Methods | Parameters $(N, \sigma, \tau)/(N, \tau)$ | False positive | False negative |
|---|---|---|---|---|
| Covariate shift | VAE-R | 5, 2, 50 | 0/50 | 0/50 |
| | | 10, 4, 40 | 0/50 | 0/50 |
| | | 20, 8, 110 | 0/50 | 0/50 |
| | VAE-R +LRP | 5, 2, 50 | 4/50 | 0/50 |
| | | 10, 4, 40 | 3/50 | 0/50 |
| | | 20, 8, 110 | 4/50 | 0/50 |
| Target shift | VAE-R | 5, 2, 50 | N/A | 13/50 |
| | | 10, 4, 40 | | 11/50 |
| | | 20, 8, 110 | | 14/50 |
| | VAE-R +LRP | 5, 2, 50 | N/A | 3/50 |
| | | 10, 4, 40 | | 2/50 |
| | | 20, 8, 110 | | 2/50 |
| Label concept shift | VAE-R | 5, 2, 50 | 0/50 | 0/50 |
| | | 10, 4, 40 | 0/50 | 0/50 |
| | | 20, 8, 110 | 0/50 | 0/50 |
| | VAE-R +LRP | 5, 2, 50 | 4/50 | 0/50 |
| | | 10, 4, 40 | 3/50 | 0/50 |
| | | 20, 8, 110 | 4/50 | 0/50 |

### D. Label concept shift

For the label concept shift, we can use a different size of the lead vehicle which never appeared in the training dataset. For testing, we collect 100 episodes with double-size leading vehicle, and one episode is shown in Fig. 7. It is reasonable to see that the predicted distance is much smaller than the ground-truth distance at the beginning of the episode since the double-size car occupies more pixels in the image than the normal-size car. The $p$-value becomes small and the detector indicates the dataset shift happens in the test episode. The method with LRP can detect such dataset shift earlier than the method without LRP since the LRP algorithm make the nonconformity measure focusing on the area which contributes more to the LEC output. Table I reports the false alarms for detection label concept shift using the VAE-based nonconformity measure with and without the LRP algorithm. The results show that our approach can detect label concept shift with a very small number of false alarms.

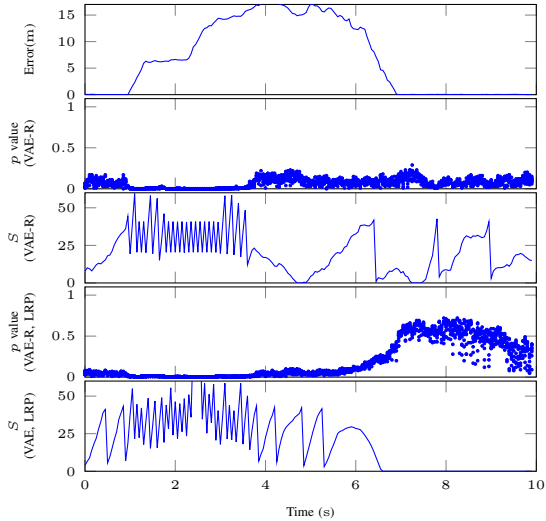

Fig. 6. Episode with dataset shift (target shift).

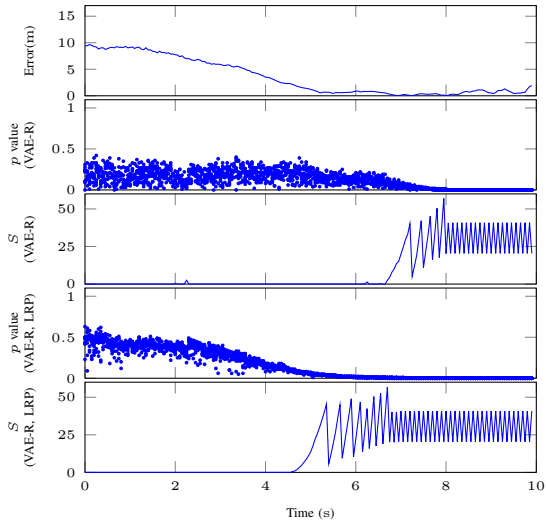

Fig. 7. Episode with dataset shift (label concept shift).

### E. Computational Efficiency

The VAE for regression model can be used for both perception and detection tasks. In order to show the real-time nature of our approach, Table II reports the minimum (min), first quartile ($Q_1$), second quartile or median ($Q_2$), third quartile ($Q_3$), and maximum (max) of the execution times of the detectors with and without the LRP algorithm for different values of $N$. From the results, see that the execution time increases as $N$ grows since the VAE for regression model needs to be run $N$ times to generate $N$ examples. The method with LRP takes a slightly longer time (about $5.5\,\text{ms}$) than the approach without LRP due to LRP computations. The execution times of the detectors are shorter than the sampling time of AEBS ($50\,\text{ms}$), and therefore, our approach is applicable for real-time dataset shifts detection.

TABLE II
EXECUTION TIMES.

|  | $N$ | min (ms) | $Q_1$ (ms) | $Q_2$ (ms) | $Q_3$ (ms) | max (ms) |
|---|---|---|---|---|---|---|
| VAE-R | 5 | 10.76 | 10.96 | 11.96 | 11.98 | 12.22 |
|  | 10 | 20.83 | 20.91 | 20.97 | 21.25 | 21.83 |
|  | 20 | 41.96 | 42.28 | 42.77 | 43.16 | 49.13 |
| VAE-R+LRP | 5 | 16.10 | 16.16 | 16.25 | 16.62 | 26.33 |
|  | 10 | 26.49 | 26.61 | 26.65 | 26.77 | 34.28 |
|  | 20 | 47.80 | 47.92 | 47.98 | 49.04 | 54.81 |

## VI. RELATED WORK

In the last decade, dataset shift has received a growing amount of interest in the computer science community. The effects of the dataset shift on the performance of models were first studied by [19]. They proposed re-weighting the observed testing samples based on the empirical training and testing probability distributions to maximize the maximum weighted log-likelihood estimate (MWLE) and to increase model performance in the presence of covariate shift. A more robust reweighing method called kernel mean matching (KMM) to treat the covariate shift was studied in [20]. Another frequently occurring shift called target shift was mentioned first by [21]. The KMM approach was adopted to target shift in classification problems, also known as label shift [22]. In [11] and [23], the label shift was analyzed in depth, and state-of-the-art approaches are proposed to minimize the impact of the shift. Label concept shift is one of the areas that is studied least since it occurs rarely and extremely difficult to solve. In literature, this shift was formalized in [24] first time.

Several anomaly detection approaches based on the conformal anomaly detection are raised, but the nonconformity measures are defined differently. Kernel density estimation (KDE) [25] and $k$-nearest neighbor [26] nonconformity measures are used for anomaly detection of the single point. For sequential anomaly detection of time trajectories, the nonconformity measures can be the sum of Hausdorff distances to $k$ nearest neighbors [27], the average of Mahalanobis distances to the $k$ nearest neighbors [28], and the sub-sequence local outlier factor [6]. However, such nonconformity measures cannot scale to the high-dimensional inputs. Recently, variational autoencoder (VAE) and deep support vector data description (deep SVDD) are used to compute the nonconformity scores for high-dimensional inputs in [7].

The purpose of LRP is to identify parts of the input that contribute most to the LEC prediction [17], and it is normally used as a tool for interpreting neural networks to help to understand if the LECs focus on the reasonable cues in the input. The propagation rules of the LRP algorithm can be applied to various neural network architectures, including convolutional networks [17], LSTMs [29], and different applications, such as text [30], image [17], and video [31]. In [32], the LRP algorithm is firstly collaborated with an anomaly detection model trying to explain the outliers in the dataset. In our approach, we incorporate the relevance computed by LRP into the nonconformity measure to improve the robustness of the detector.

## VII. Conclusions

In this paper, we discuss formal definitions and provide some practical examples for a variety of dataset shifts in learning-enabled CPS. Focusing on these dataset shifts, we propose a detection approach based on inductive conformal anomaly detection. A VAE for regression model is trained to compute the nonconformity of new test inputs with respect to the training data, which enables the detection to take both inputs and outputs into consideration. Moreover, the layer-wise relevance propagation (LRP) algorithm is incorporated to improve the robustness of the detection. A simulation case study of an advanced emergency braking system (AEBS) is utilized to evaluate our approach. Experimental results demonstrate our approach can detect different types of dataset shifts with small false alarms, and the execution time is relatively short enabling real-time detection. Although our approach takes the output into account, the results do not show any significant improvements compared to the methods which only use the input. Extending to this work, we should design another experiment to demonstrate the benefit of taking the output into consideration. Incorporating the attention techniques into the detector is another possible direction for future work.